\documentclass[10pt,twocolumn,letterpaper]{article}

\usepackage[pagenumbers]{cvpr} 

\usepackage{dsfont}
\usepackage[accsupp]{axessibility}

\usepackage{microtype}

\renewcommand{\paragraph}[1]{\vspace{.5em}\noindent\textbf{#1.}}

\usepackage{soul}
\setuldepth{foobar}

\usepackage{multirow}
\usepackage{colortbl}
\definecolor{verylightgray}{gray}{0.9}

\definecolor{cvprblue}{rgb}{0.21,0.49,0.74}
\usepackage[pagebackref,breaklinks,colorlinks,allcolors=cvprblue]{hyperref}

\def\paperID{2} 
\def\confName{CVPR}
\def\confYear{2026}

\title{Positive-First Most Ambiguous: A Simple Active Learning Criterion for Interactive Retrieval of Rare Categories}

\author{
Kawtar Zaher$^{1,2}$\\{\tt\small kawtar.zaher@inria.fr}
\and
Olivier Buisson$^{2}$\\{\tt\small obuisson@ina.fr}
\and
Alexis Joly$^{1}$\\{\tt\small alexis.joly@inria.fr}\\
\and
$^{1}$INRIA, LIRMM, Université de Montpellier, France\\
$^{2}$Institut National de l'Audiovisuel, France
}

\begin{document}
\maketitle
\begin{abstract}
    Real-world fine-grained visual retrieval often requires discovering a rare concept from large unlabeled collections with minimal supervision. This is especially critical in biodiversity monitoring, ecological studies, and long-tailed visual domains, where the target may represent only a tiny fraction of the data, creating highly imbalanced binary problems. Interactive retrieval with relevance feedback offers a practical solution: starting from a small query, the system selects candidates for binary user annotation and iteratively refines a lightweight classifier. While Active Learning (AL) is commonly used to guide selection, conventional AL assumes symmetric class priors and large annotation budgets, limiting effectiveness in imbalanced, low-budget, low-latency settings. We introduce Positive-First Most Ambiguous (PF-MA), a simple yet effective AL criterion that explicitly addresses the class imbalance asymmetry: it prioritizes near-boundary samples while favoring likely positives, enabling rapid discovery of subtle visual categories while maintaining informativeness. Unlike standard methods that oversample negatives, PF-MA consistently returns small batches with a high proportion of relevant samples, improving early retrieval and user satisfaction. To capture retrieval diversity, we also propose a class coverage metric that measures how well selected positives span the visual variability of the target class. Experiments on long-tailed datasets, including fine-grained botanical data, demonstrate that PF-MA consistently outperforms strong baselines in both coverage and classifier performance, across varying class sizes and descriptors. Our results highlight that aligning AL with the asymmetric and user-centric objectives of interactive fine-grained retrieval enables simple yet powerful solutions for retrieving rare and visually subtle categories in realistic human-in-the-loop settings.
\end{abstract}    
\section{Introduction}
\label{introduction}

Fine-grained visual discovery in large-scale collections is increasingly important: in real-world scenarios, users seek a specific and visually subtle concept (e.g., a rare species, an object variant, or an unusual visual context) hidden among millions of irrelevant items. This typically induces a severely imbalanced setting where relevant instances may occur at extremely low frequencies (e.g., 1 in 1000 or 1 in 10000).

\begin{figure}[h]
    \centering
    \includegraphics[width=\linewidth]{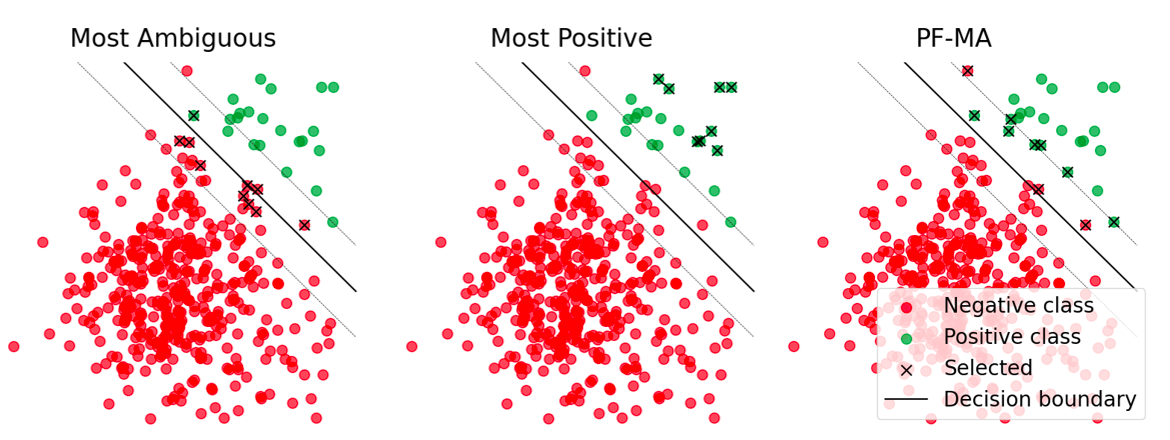}
    \caption{Comparison of selected samples. \textit{MA} (left): near-boundary negatives are oversampled. \textit{MP} (middle): only positives far from the boundary are selected. \textit{PF-MA} (right): balance between relevant positives and negatives around the boundary.}
    \label{fig:selection-strategies}
\end{figure}

\begin{figure}[h]
    \centering
    \includegraphics[width=1\linewidth]{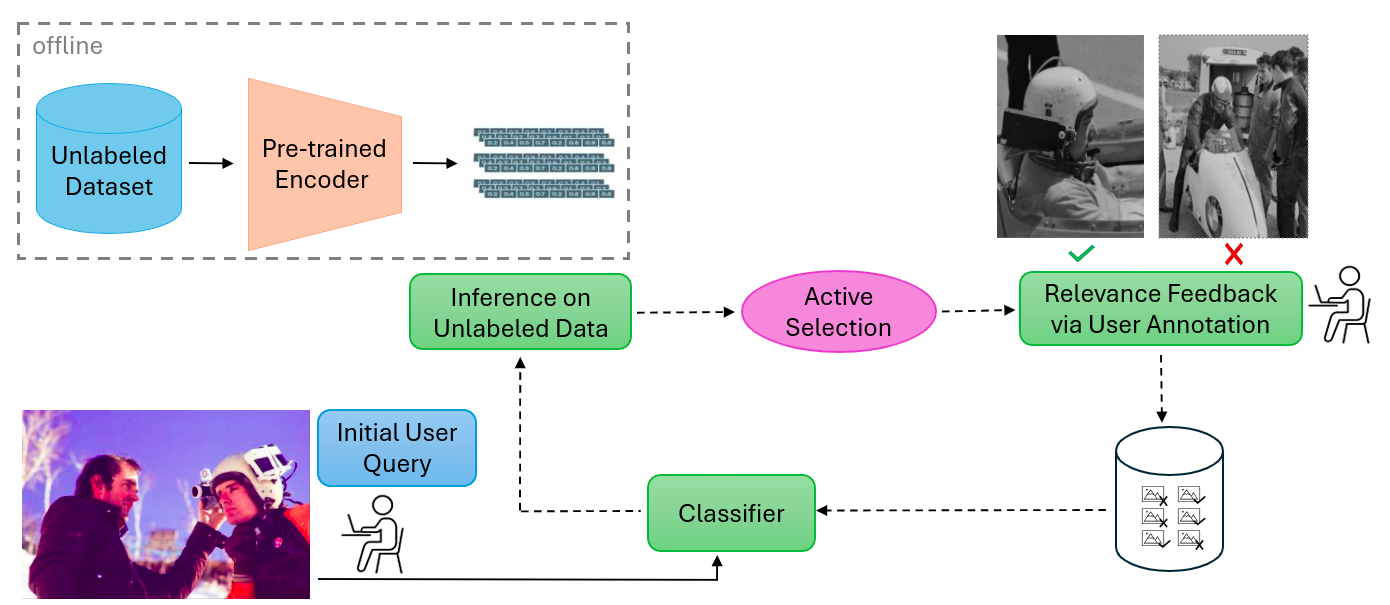}
    \caption{Interactive retrieval process.}
    \label{fig:process}
\end{figure}

We consider interactive retrieval, where the system progressively constructs a novel user-defined class-of-interest using minimal supervision. The overall interaction loop is shown in Fig.~\ref{fig:process}: starting from a small initial query, the system proposes candidate images, the user provides binary Relevance Feedback (RF)~\cite{ngo2016image}, and a lightweight binary classifier refines the target concept over iterations based on the user feedback. The objective is not to train a general-purpose model, but to quickly retrieve relevant and diverse instances of the user-defined class under strict constraints: 
\begin{enumerate}[label=(\roman*)]
    \item severe class imbalance (rare positives vs. abundant negatives),
    \item tiny per-iteration annotation budgets,
    \item and low latency between iterations (few seconds).
\end{enumerate}
Constraints (ii) and (iii) reflect human limitations and user fatigue: users can only label a few samples and wait a few seconds for the next batch.

Active Learning (AL)~\cite{settles2009active} naturally guides sample selection, but conventional AL assumes symmetric class priors and labeling utility, aiming to improve global classification by answering \emph{“which samples should be labeled to train the best overall model?”}. 

In interactive fine-grained retrieval, sample utility is inherently asymmetric: querying a predicted-positive sample provides substantially higher expected utility than querying a predicted-negative one, as it instantly improves user satisfaction, particularly under tight annotation budgets. Standard strategies ignore this asymmetry, wasting queries on ambiguous negatives. This creates a tension between \emph{informativeness} and \emph{relevance}: uncertainty-focused strategies sample near the decision boundary, naturally crowded with negatives, while confidence-based selection often returns redundant positives with limited training value, as illustrated in Fig.~\ref{fig:selection-strategies}. The key question becomes: \emph{"which samples should be labeled to rapidly and broadly characterize the rare class-of-interest?"}. The goal is not to improve general classification, but to rapidly and effectively construct the user's class-of-interest.

To tackle this problem, we introduce \textbf{Positive-First Most Ambiguous (PF-MA)}, a simple AL criterion designed for interactive class retrieval under severe class imbalance and real-world constraints. PF-MA prioritizes ambiguous samples while explicitly favoring likely positives, enabling rapid rare class discovery without sacrificing boundary refinement, and adapting to the asymmetric nature of our problem. This equilibrium, shown in Fig.~\ref{fig:selection-strategies}, ensures that the system consistently presents the user with relevant examples, enhancing early satisfaction and visual confirmation, while also exposing the model to hard negatives that are crucial for learning precise decision boundaries. 

Moreover, interactive retrieval also requires \emph{diversity}: users benefit from covering different visual modes of the class (e.g., pose, context, appearance), not only from retrieving more redundant positives. We therefore propose a \textbf{class coverage metric} that evaluates retrieval quality beyond the number of discovered positives, measuring how broadly the retrieved set spans the underlying class manifold.

We demonstrate the effectiveness of \textit{PF-MA} across long-tailed datasets, including fine-grained images, using modern pretrained visual descriptors. Beyond the intrinsic rarity of the user-defined class in our highly imbalanced setting, we leverage long-tailed datasets to model the unknown and variable size of this class, ranging from a handful of instances to several hundreds. Despite its simplicity, PF-MA consistently improves early-stage retrieval coverage and classifier quality across classes of varying sizes. Our results highlight the practical benefits for human-in-the-loop fine-grained retrieval: high positive ratios, diverse retrieval, and rapid user satisfaction with minimal supervision.
\section{Related Work}
\label{relatedwork}

\paragraph{Interactive Retrieval and Relevance Feedback}
Interactive retrieval aims to construct a user-defined class with minimal supervision through an iterative process. Unlike Novel Class Discovery~\cite{troisemaine2023novel}, which identifies all novel categories, interactive retrieval focuses on retrieving a single concept defined by the user. It is related to Few-Shot Learning (FSL)~\cite{song2023comprehensive} which learns from limited examples, but FSL assumes a fixed label space and optimizes overall classification accuracy. In contrast, interactive retrieval operates without predefined categories and emphasizes retrieving diverse and relevant instances.

Content-Based Image Retrieval (CBIR)~\cite{long2003fundamentals, hameed2021content, srivastava2023content} is closely related, yet often targets instance-level similarity. Early systems relied on handcrafted features and relevance feedback (RF) to iteratively refine queries~\cite{rocchio1971relevance, cox2000bayesian, ngo2016image}. With deep learning, learned embeddings significantly improved retrieval~\cite{babenko2014neural}. Recent works explore interactive learning with deep embeddings and user-in-the-loop refinement~\cite{tzelepi2016relevance}. However, these similarity-driven strategies frequently produce near-duplicate suggestions, limiting diversity and training signal.

\paragraph{Human-in-the-Loop, Interactive Learning and Active Learning}
Human-in-the-loop learning incorporates user feedback to refine model predictions~\cite{monarch2021human}. To address the near-duplicate limitation, Active Learning (AL)~\cite{settles2009active} selects informative samples to reduce annotation effort. Classic AL focuses on improving model performance with minimal labeling using uncertainty~\cite{lewis1994sequential, scheffer2001active, tong2001support}, diversity~\cite{brinker2003incorporating, xu2007incorporating}, or representativeness strategies such as CoreSet methods~\cite{sener2017active} and unlabeled data modeling~\cite{sener2017active, gissin2019discriminative}. These approaches are primarily designed for supervised classification with fixed label spaces and balanced datasets, and typically assume symmetric labeling utility across classes, disregarding the class-of-interest, while optimizing global accuracy. Recent deep AL methods~\cite{li2024survey} further rely on repeated retraining or stochastic forward passes, introducing additional computational overhead, unfeasible in our low latency setting.

In retrieval settings, AL has been explored to guide interactive feedback toward relevant results~\cite{gosselin2008active, barz2018information, bar2024active}, but often requires repeated retraining or scoring over large candidate pools at each iteration, which conflicts with the strict latency constraints.

In biodiversity applications, interactive systems have been used to support species identification and knowledge acquisition. Platforms such as Pl@ntNet leverage user feedback and collective intelligence to improve recognition models~\cite{joly2016look}. However, most systems focus on classification rather than retrieval of flexible, user-defined visual patterns.

\paragraph{Class Imbalance in Active Learning}
Interactive retrieval naturally induces extreme class imbalance, where the class-of-interest represents only a small fraction of the dataset. Class-agnostic AL methods~\cite{attenberg2013class, sener2017active, gissin2019discriminative} tend to oversample majority regions and neglect rare relevant samples. Imbalance-aware variants~\cite{aggarwal2020active, aggarwal2021minority, bengar2022class, fairstein2024class} rebalance classes using frequency estimates from the labeled set, which are unreliable under very small annotation budgets, while~\cite{kothawade2021similar} assumes prior knowledge of the rare class. Importantly, beyond statistical imbalance, interactive retrieval exhibits an asymmetric labeling utility: querying likely positives has higher value than querying negatives. Our objective is therefore not distribution matching, but sustained and rapid discovery of positives.

\paragraph{Fine-Grained Recognition and Visual Similarity}
Fine-grained visual categorization (FGVC) focuses on distinguishing visually similar categories, such as species or object subtypes~\cite{wei2019deep}, using part-based models, attention, or metric learning to capture subtle differences~\cite{peng2017object, qian2015fine}. These approaches improve retrieval and classification but typically require labeled training data and predefined classes. In real-world scenarios such as biodiversity monitoring, users often search for concepts that do not exactly match existing categories. While recent works have investigated open-set and open-world recognition to address this limitation~\cite{bendale2016towards, joseph2021towards}, they do not explicitly support iterative, user-guided refinement. Interactive retrieval complements these approaches by enabling users to progressively define visual concepts.

\paragraph{Summary}
Overall, prior work highlights the importance of interactive feedback, active learning and fine-grained visual understanding. However, existing methods either rely on predefined categories, optimize global accuracy under symmetric utility assumptions, depend on reliable class-frequency estimates, or lack iterative refinement under strict latency constraints. Our work explicitly addresses the asymmetric utility and extreme imbalance of interactive fine-grained retrieval, combining deep embeddings, interactive feedback, and scalable selection to support rapid discovery of user-defined concepts in long-tailed settings.
\section{Methodology}
\label{methodology}

In this section, we first formulate our setup before presenting our novel \textbf{Positive-First Most Ambiguous (PF-MA)} Active Learning selection criterion.

\subsection{Problem Formulation}
Our data is a set of $N$ unlabeled images $\{I_i\}_{i \in [1,N]}$, encoded using a pretrained neural network $\Phi$ to obtain the initial dataset $D = \{x_i\}_{i \in [1,N]}$, where $x_i = \Phi(I_i) \in \mathds{R}^d$.
To retrieve the class-of-interest, the user provides a first query containing $N_p$ positive images, i.e. that contain the desired concept, and $N_n$ negative images, i.e. different than the desired concept. In this study we consider very small values of $N_p$ and $N_n$ ($N_p=1$ and $N_n=5$ in our experiments) corresponding to a very limited annotation budget scenario. The provided query is used to initiate a labeled training set $D_l$, where positive images are labeled $1$ and negative images are labeled $0$. 
\\
\\
The learning and retrieval loop (Fig.~\ref{fig:process}) is then performed for $T$ iterations. In our experiments, we consider values of $T$ up to $T=25$ iterations. Each iteration $1 \leq t \leq T$ is conducted as follows:
\begin{enumerate}
    \item A classifier $f : x_i \in \mathds{R}^d \mapsto [0, 1]$ is trained on $D_l$ to recognize the class-of-interest, and classify each image as positive, i.e. in the class-of-interest, or negative, i.e. outside the class-of-interest.
    \item An AL strategy is used to score each sample in $D \smallsetminus D_l$.
    \item Given an annotation budget $b$, the top $b$ samples with the highest AL scores are selected to obtain the selected set $S_t$.
    \item The selected samples are annotated by the user and added to the training data: $D_l \leftarrow D_l \cup S_t$. \\
\end{enumerate}

\noindent As the loop is performed in a interactive manner, where the user labels the selected samples, not only the wait time of the user before the next set of selected samples should be realistic, of the order of a few seconds, but also, the amount of selected samples should be easily handled by the user. These constraints impose the use of a light model for the classifier, as we are in interactive, low latency, low annotation budget setting. Like the work of \cite{aggarwal2022optimizing} that showed that training a lightweight classifier outperforms the finetuning of the encoder model, we use a \textit{linear SVM} for its speed in low-data regimes. However, we use a much smaller budget $b = 10$, where they use $b=200$.

\subsection{Active Sample Selection and Limitations}

As mentioned before, our selection criterion should not only \textit{train a good classifier}, but also ensure that \textit{each iteration adds a diverse set of positive images covering the class-of-interest}. Two baseline strategies arise:

\begin{itemize}
    \item An uncertainty criterion that provides maximum informativeness to the classifier. Because of the low latency constraint, we use the simplest uncertainty criterion called \textit{MA} for \textit{Most Ambiguous}. The score for each sample $x_i$ is:
    \begin{equation}
    MA(x_i) = 1 - | 0.5 - f(x_i) |
    \end{equation}
    \textit{MA} ensures a strong classifier but fails to respect the rapid class filling requirement: under severe class imbalance, near-boundary points are mostly negatives (Fig.~\ref{fig:selection-strategies}). This requirement is of crucial importance, since the human user may decide to stop before finishing the process. As a result, relevant samples should be presented since the start.
    \item A criterion that ensures the selection of positive samples. This strategy is called \textit{MP} for \textit{Most Positive}:
    \begin{equation}
    MP(x_i) = f(x_i)
    \end{equation}
    \textit{MP} prioritizes sample utility for the user through fast retrieval, but these samples are often redundant and belong to the already explored subspaces, offering limited information to the classifier.
\end{itemize}
Fig. \ref{fig:teaser} (top and middle) illustrates these trade-offs: \textit{MA} provides informative but mostly negative samples, while \textit{MP} rapidly retrieves positives with limited diversity. \\
After outlining the limitations of traditional strategies, we introduce our \textit{PF-MA} criterion, which aims to reconcile informative diversity with the rapid retrieval of positives.

\subsection{Positive-First Most Ambiguous AL Strategy}
Our \textit{PF-MA} strategy combines the strengths of uncertainty (\textit{MA}) and confidence (\textit{MP}) selection (Fig.~\ref{fig:selection-strategies}). The goal is to prioritize samples that are likely positive, maximizing immediate user satisfaction, while still maintaining sufficient ambiguity to provide informative signals to the classifier. Formally, the \textit{PF-MA} score is:
\begin{equation}
\begin{aligned}
\text{\textit{PF-MA}}(x_i) = (1 - | 0.5 - f(x_i) |) \times \mathds{1}_{(f(x_i) \geq 0.5)} \\
+ f(x_i) \times \mathds{1}_{(f(x_i) < 0.5)}
\end{aligned}
\end{equation}
Intuitively, \textit{PF-MA} asymmetrically prioritizes the positive half-space to correct the symmetry bias of \textit{MA}. It first selects positives near the decision boundary, then less ambiguous positives, and only includes negatives if the iteration budget is not yet filled. This strategy ensures that early positive samples provide high immediate user satisfaction, while still exposing the classifier to informative samples for improved boundary refinement.

\subsection{Assessing the diversity: Class Coverage Metric}

A straightforward metric to assess retrieval is the \textbf{proportion of selected positives}. For a class-of-interest $C$ of size $k_C$, let $P_t$ be the set of positives returned up to iteration $t$, then: $pos_t^C = \# P_t / k_C$. While easy to compute, this metric does not capture diversity. Consider searching for images of birds: retrieving 30 images all showing birds perched on similar branches provides less value than retrieving 30 images showing different behaviors like flying, hunting, nesting, perched on rocks. This metric treats both scenarios equally, even though the latter is far more informative for the user. Retrieval metrics like mean Average Precision (mAP) also primarily focus on the number of positives.

Some diversity metrics rely on pairwise distances between samples. But in our case, classes can be naturally concentrated or spread across multiple visual modes. Low pairwise distances in a compact class do not necessarily indicate low diversity, as different poses or contexts may be clustered tightly in feature space. Conversely, classes with spread-out samples might show high pairwise distances even if only a few modes are captured. Other metrics, like the Vendi score~\cite{friedman2022vendi}, measure diversity as the entropy of feature variances, but these approaches ignore the actual topology of the underlying class. In other words, a subset could score high simply because it spans unusual directions in the feature space, without truly representing the visual modes of the reference set that matter to the user.

In interactive retrieval, we mainly care that the model queries the user across all regions of the class-of-interest. To capture this, we propose a \textbf{class coverage metric} $cov_t^C$, which measure how well the retrieved positives span the different visual modes of the class. For each class, we perform offline \textit{K-means} clustering to obtain $K$ clusters $\{CLS_i\}_{i=1}^{K}$ (the labels are only used for evaluation). Each retrieved positive is assigned to a cluster, and coverage is the fraction of clusters with at least one positive:
\begin{equation}
cov_t^C = \frac{\#\{ CLS_i \mid \exists x \in P_t, x \in CLS_i \}}{K}.
\end{equation}
This metric rewards the system for retrieving across the whole class rather than from a single visual mode. We choose $K=32$ and justify this choice through experiments. For more stable estimates, we average over 10 K-means runs. For classes smaller than $K$, each sample is treated as its own cluster.

\section{Experimental Framework}


\subsection{Datasets}
Real world fine-grained datasets frequently suffer from skewed class proportions, where classes have varying sizes and thus varying frequencies. This often results in classifiers having better performances regarding majority classes, whereas minority classes are overlooked. The problem is that commonly used vision datasets are artificially well-balanced. We rather focus on datasets with long-tailed class size distributions, where a few classes account for the majority of the dataset, and a larger share of classes represent a very small number of samples. We make use of such datasets as their imbalance presents a great proxy for the unknown size of the class-of-the interest. In other words, we simulate scenarios where the users might search for either common patterns (large classes) or rare concepts (small classes), without knowing in advance which case applies. We use the datasets \textbf{Cifar100-LT} \cite{cao2019learning}, \textbf{ImageNet-LT} \cite{liu2019large}, and \textbf{PlantNet300K} \cite{garcin2021pl}. We report the statistics of class frequencies of these datasets in Tab.~\ref{tab:freqs}, highlighting not only the overall rarity of the researched concepts but also the varying degrees of that rarity.

\begin{table}[h]
    \centering
    \caption{Statistics of class frequencies per dataset.}
    \label{tab:freqs}
    \resizebox{0.9\columnwidth}{!}{
    \begin{tabular}{|c|cccc|}
    \hline
    dataset & min & max & mean & median \\
    \hline
    \hline
    Cifar100-LT & $0.079\%$ & $3.97\%$ & $1\%$ & $0.56\%$ \\
    ImageNet-LT & $0.004\%$ & $1.1\%$ & $0.1\%$ & $0.06\%$ \\
    PlantNet300K & $0.001\%$ & $2.96\%$ & $0.09\%$ & $0.01\%$ \\
    \hline
    \end{tabular}
    }
\end{table}

\subsection{Descriptor Models} 
We use two pre-trained models to describe our images, both of them based on the \textit{ViT-L14} \cite{dosovitskiy2020image} architecture, but trained via two different frameworks: \textbf{CLIP} \cite{radford2021learning} and  \textbf{DINOv2} \cite{oquab2023dinov2}, as they showcase great performances for other tasks. We specifically selected these two models due to their strong performance in few-shot retrieval scenarios. In particular, DINOv2 excels at capturing fine-grained visual similarity, making it especially effective for purely image-based retrieval tasks. The main difference between both frameworks is that DINOv2 model is pre-trained using self-supervision (on images only), whereas CLIP is pre-trained using contrastive learning on images and associated textual descriptions (language-supervised). In addition, CLIP could allow for the future use of textual queries.

\subsection{Initial Query} 
For the retrieval, we randomly select $Q$ queries per class, where each query contains $N_p$ positive image and $N_n$ negative images. Generally, $N_p \leq N_n$ as the user can initially provide very few positive images, compared to negative ones. We choose $Q=10$ different initial queries per class to ensure robust evaluation across different starting points. We set $N_p=1$ and $N_n=5$.

\subsection{Baselines} 
To rigorously benchmark our method, we compare it against a range of baselines, from simple heuristics to advanced state-of-the-art AL strategies, as described below:
\begin{itemize}
    \item \textit{Random}: Random selection.
    \item \textit{MA}: Most Ambiguous.
    \item \textit{MP}: Most Positive.
    \item \textit{DAL}: Discriminative Active Learning \cite{gissin2019discriminative}. \textit{DAL} chooses the samples that are more similar to the unlabeled data than to the labeled data, using a labeled vs. unlabeled classifier.
    \item \textit{CoreSet}: CoreSet selection~\cite{sener2017active}.
    \item \textit{ALAMP:} We choose \textit{ALAMP} \cite{aggarwal2022optimizing} because of its close settings to our problem, i.e. a small budget and a linear classifier. In brief, \textit{ALAMP} assigns scores to samples at iteration $t$ as follows:
    \begin{equation}
        ALAMP(x_i) = \frac{marg_{t-1}(x_i) - marg_{t}(x_i)}{marg_{t-1}(x_i) + marg_{t}(x_i)}
    \end{equation}
    where $marg_t(x_i)$ is the margin sampling score. The margin sampling score is defined as the difference between the probabilities of the top-2 predicted classes. The larger the margin score, the further the probabilities assigned, and the more certain the sample is. As a result, \textit{ALAMP} prioritizes samples that switched from high certainty to high uncertainty. When the difference in uncerainty is the same, the denominator prioritizes samples with overall low certainty.
\end{itemize}
We also combine the uncertainty strategies with diversity methods. We choose straightforward diversity terms, to keep the computation times low:
\begin{itemize}
    \item \textit{*-S} (for step): where * $\in$ \{\textit{MA}, \textit{MP}\}. The samples are ensured to be diverse by selecting one sample each $S=5$ samples from the set of ordered samples.
    \item \textit{*-D} (for distance): where * $\in$ \{\textit{MA}, \textit{MP}\}. A larger number $B=50$ of samples is selected using the uncertainty criterion to form a pool. Then the $b$ samples are selected from this pool by iteratively sampling the farthest sample from the already selected ones.
\end{itemize}

\subsection{Annotation Budget}
Because annotation efficiency is central in interactive scenarios, we clarify here the annotation budget constraints imposed at each iteration and the rationale behind these choices. \cite{aggarwal2022optimizing} clearly mentions that small budgets ($200$ annotations) present a challenge for AL. In our case, the annotations are conducted by the user, making it impossible to annotate hundreds of samples. Therefore, we further restrict the annotation budget to merely $b=10$ samples at each iteration, for different number of iterations ranging from $T=1$ to $T=25$ iterations. In this work, ground truth labels are used as a proxy for user annotations.

\subsection{Additional Metrics}
While our primary evaluation metric is class coverage, which directly reflects the effectiveness of the retrieval task, we may also use other metrics. The \textbf{f1-score} on a held-out test set helps assess the performance of the underlying binary classifier and its generalization ability. The \textbf{proportion of selected positives} disregards the quality of the retrieved samples. In contrast, only the \textbf{class coverage} is able to fully capture the actual goal of the retrieval task: the ability of the selection strategies to retrieve positive samples and the diversity of these samples. Therefore, although we may include these other metrics, our analysis and conclusions are primarily based on the class coverage. 
\section{Experimental Results}

\subsection{Inadequacy of Representativeness-based Active Learning Strategies}
To motivate our approach, we begin with a preliminary experiment highlighting the limitations of widely used AL methods that aim to reproduce the unlabeled data distribution, as previously explained. These representativeness-based strategies are commonly employed but they fail in imbalanced settings where minority classes are of primary interest. In Tab.~\ref{tab:coreset}, we compare our uncertainty criterion \textit{MA} with \textit{Rand}, \textit{CoreSet} and \textit{DAL} on CIFAR100-LT, using DINOv2. More results across iterations are presented in App.~\ref{app:generalization_f1}.
\begin{table}[h]
    \centering
    \caption{Performance comparison of representativeness-based AL methods vs. uncertainty-based selection on CIFAR100-LT at iteration $25$.}
    \label{tab:coreset}
    \resizebox{0.7\columnwidth}{!}{
    \begin{tabular}{|c|cccc|}
    \hline
    metric & \textit{Rand} & \textit{DAL} & \textit{CoreSet} & \textit{MA} \\
    \hline
    \hline
    $cov_{25}$ & 0.111 & 0.096 & 0.087 & \textbf{0.899} \\
    $f1_{25}$ & 0.381 & 0.321 & 0.164 & \textbf{0.857} \\
    $pos_{25}$ & 0.036 & 0.032 & 0.035 & \textbf{0.61} \\
    \hline
    \end{tabular}
    }
\end{table}
The results clearly demonstrate the inadequacy of representativeness methods for our interactive search problem. \textit{CoreSet} and \textit{DAL} achieve coverage below $0.1$ and discover fewer than $4\%$ of the class. This demonstrates that distribution-based methods systematically oversample from majority regions, which in long-tailed settings contain mostly irrelevant samples. The poor performance of these methods confirms our hypothesis that reproducing the data distribution is fundamentally unsuited for interactive retrieval, where the goal is to maximize discovery of the minority class-of-interest. While random selection is also not adapted to our setting, \textit{MA} provides a valid comparison baseline. Consequently, we focus our investigation on uncertainty-based strategies, which show greater potential in addressing these challenges.

\subsection{Performance Across Multiple Datasets and Feature Descriptors}

In this section, having established the limitations of representativeness-based methods, we compare our selection criterion to uncertainty and diversity-based methods, as they provide strong comparison baselines, per the previous section. Specifically, we assess performance across diverse datasets and feature descriptors to determine the robustness of our approach. Tab.~\ref{tab:generalization_imbalanced} demonstrates the consistent superiority of \textit{PF-MA} across three imbalanced datasets and two feature descriptors: our method achieves the highest coverage scores in nearly all settings. We report results across more iterations in App.~\ref{app:generalization_cov}, and the classifier performance results in App.~\ref{app:generalization_f1}.

\begin{table}[h!]
    \centering
    \caption{Class coverage scores $cov_{25}$ at iteration $25$ for different AL methods, datasets and descriptors.}
    \label{tab:generalization_imbalanced}
    \resizebox{1\columnwidth}{!}{
    \begin{tabular}{|c|cc|cc|cc|}
    \hline
    \multirow{ 2}{*}{method} & \multicolumn{2}{c|}{Cifar100-LT} & \multicolumn{2}{c|}{ImageNet-LT} & \multicolumn{2}{c|}{PlantNet300K}\\ 
    & CLIP & DINOv2 & CLIP & DINOv2 & CLIP & DINOv2 \\
    \hline
    \hline
    \textit{MA} & 0.89 & 0.899 & 0.852 & 0.804 & \textbf{0.596} & 0.678 \\
    \textit{MP} & 0.875 & 0.877 & 0.821 & 0.776 & 0.565 & 0.628 \\
    \textit{ALAMP} & 0.811 & 0.892 & 0.744 & 0.771 & 0.406 & 0.605 \\
    \textit{MA-S} & 0.88 & 0.895 & 0.844 & 0.802 & 0.583 & 0.671 \\
    \textit{MA-D} & 0.903 & 0.84 & 0.842 & 0.751 & 0.595 & 0.679 \\
    \textit{MP-S} & 0.869 & 0.882 & 0.815 & 0.777 & 0.552 & 0.625 \\
    \textit{MP-D} & 0.881 & 0.86 & 0.816 & 0.76 & 0.563 & 0.639 \\
    \rowcolor{verylightgray} \textit{PF-MA} & \textbf{0.908} & \textbf{0.954} & \textbf{0.861} & \textbf{0.844} & \textbf{0.596} & \textbf{0.684} \\
    \hline
    \end{tabular}
    }
\end{table} 
Among uncertainty-based methods, \textit{PF-MA} outperforms both \textit{MA} and \textit{MP} baselines, with the advantage becoming more pronounced over iterations. This consistent improvement demonstrates that explicitly prioritizing likely positives accelerates class discovery compared to pure uncertainty sampling. The diversity-enhanced variants show unstable results, as they can underperform or overperfom their base methods, depending on the dataset and the used descriptor. For \textit{PF-MA}, DINOv2 generally achieves higher coverage than CLIP, due to the intrinsic strength of its visual features in purely image-based retrieval scenarios. However, \textit{PF-MA} maintains its advantage across both descriptors, indicating that the method's effectiveness is not dependent on a specific feature representation. The consistent performance across three diverse long-tailed datasets, from low resolution images (Cifar100-LT) to large scale natural images (ImageNet-LT) to fine-grained specialized botanical data (PlantNet300K), demonstrates the generalizability of \textit{PF-MA} for long-tailed and fine-grained interactive retrieval tasks.

\begin{figure*}[h!]
    \centering
    \includegraphics[width=1.45\columnwidth]{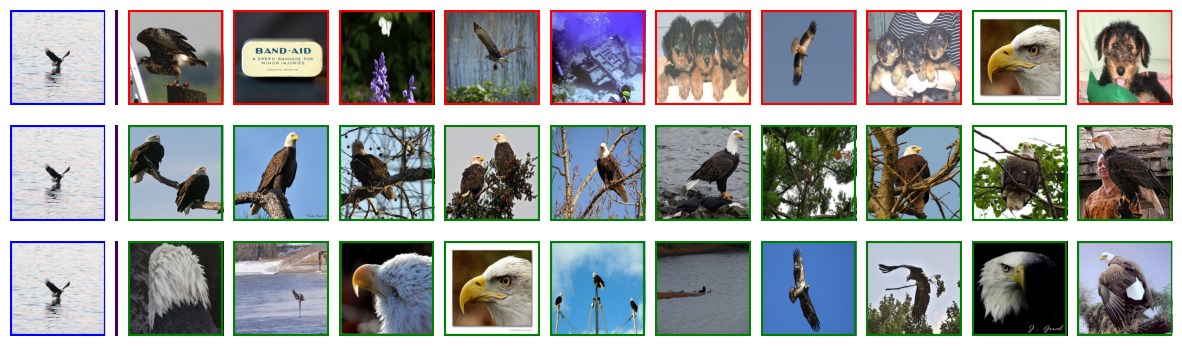}
    \caption{Retrieval results for a bird species at iteration $5$ ($50$ total annotations). The initial query image is in blue, the positive images in green and the negative ones in red. Top: \textit{MA}. Middle: \textit{MP}. Bottom: \textit{PF-MA}}
    \label{fig:teaser}
\end{figure*}

\subsection{Influence of Coverage Granularity}
To better understand the behavior of our method under varying clustering resolutions, we investigate the influence of coverage granularity. Specifically, we examine how the number of clusters $K$, used to define class coverage, affects performance. We report in Tab.~\ref{tab:granularity} the class coverage scores at iteration $25$ for different values of $K \in \{16, 32, 64\}$ and across different iteration in App.~\ref{app:coverage}.
\begin{table}[h!]
    \centering
    \caption{Class coverage scores $cov_{25}$ at iteration $25$ for different $K$ values. Second best is underlined.}
    \label{tab:granularity}
    \resizebox{0.55\columnwidth}{!}{
    \begin{tabular}{|c|ccc|}
    \hline
    \multirow{ 2}{*}{method} & \multicolumn{3}{c|}{$K$} \\ 
    & 16 & 32 & 64 \\
    \hline
    \hline
    \textit{MA} & \underline{0.848} & \underline{0.804} & 0.738 \\
    \textit{MP} & 0.776 & 0.776 & \underline{0.764} \\
    \textit{ALAMP} & 0.808 & 0.771 & 0.712 \\
    \textit{MA-S} & \underline{0.848} & 0.802 & 0.731 \\
    \textit{MA-D} & 0.833 & 0.751 & 0.648 \\
    \textit{MP-S} & 0.779 & 0.777 & \underline{0.764} \\
    \textit{MP-D} & 0.773 & 0.76 & 0.723 \\
    \rowcolor{verylightgray} \textit{PF-MA} & \textbf{0.849} & \textbf{0.844} & \textbf{0.843} \\
    \hline
    \end{tabular}
    }
\end{table}
These results, obtained on ImageNet-LT, using DINOv2 features, reveal two key observations. First, our method \textit{PF-MA} consistently outperforms other baselines across all $K$ values and iterations, demonstrating its ability to well cover the class-of-interest whatever the visual granularity considered (i.e. large sub-classes or small sub-classes). In contrast, the performance of other methods varies with $K$, and no single baseline maintains a consistent ranking. This shows their sensitivity to the choice of the visual granularity.
We also observe that smaller $K$ values yield higher coverage scores, as it becomes easier to populate more clusters. However, this may reduce the metric's distinctive power. On the other hand, larger $K$ values lead to more meaningful and differentiated scores, but fewer classes meet the minimum cluster size.To balance these trade-offs, we enforce a constraint that at least $50\%$ of classes must be eligible for clustering. we choose $K=32$, while noting that \textit{PF-MA} remains superior regardless of the $K$ value.

\subsection{Qualitative Results}
In Fig.~\ref{fig:teaser}, we show the visual retrieval results, provided the first query in blue, at iteration $5$. It is clear that \textit{MA} provides a large number of irrelevant images, reducing the user satisfaction. \textit{MP} selects only positive images, but the selected samples are very similar, with centered instances with nearly identical backgrounds. In contrast, \textit{PF-MA} provides positive images with diverse backgrounds and object positions.

\subsection{Performance Factors: Search Iterations and Class Size} 
To further dissect what drives performance, we turn our attention to two key factors: the number of search iterations and the size of the class-of-interest. These analyses provide insight into the scalability and efficiency of our approach in different retrieval scenarios.
We analyze the class coverage results with regard to these two factors, using DINOv2 features. Supplementary figures are presented in App.~\ref{app:iteration_class_size}. We also report the classifier performance results in~\ref{app:iteration_class_size_f1}.
\subsubsection{Varying the Search Iteration}
\begin{figure}[h!]
    \centering
    \begin{subfigure}[t]{\columnwidth}
        \centering
        \includegraphics[width=\linewidth]{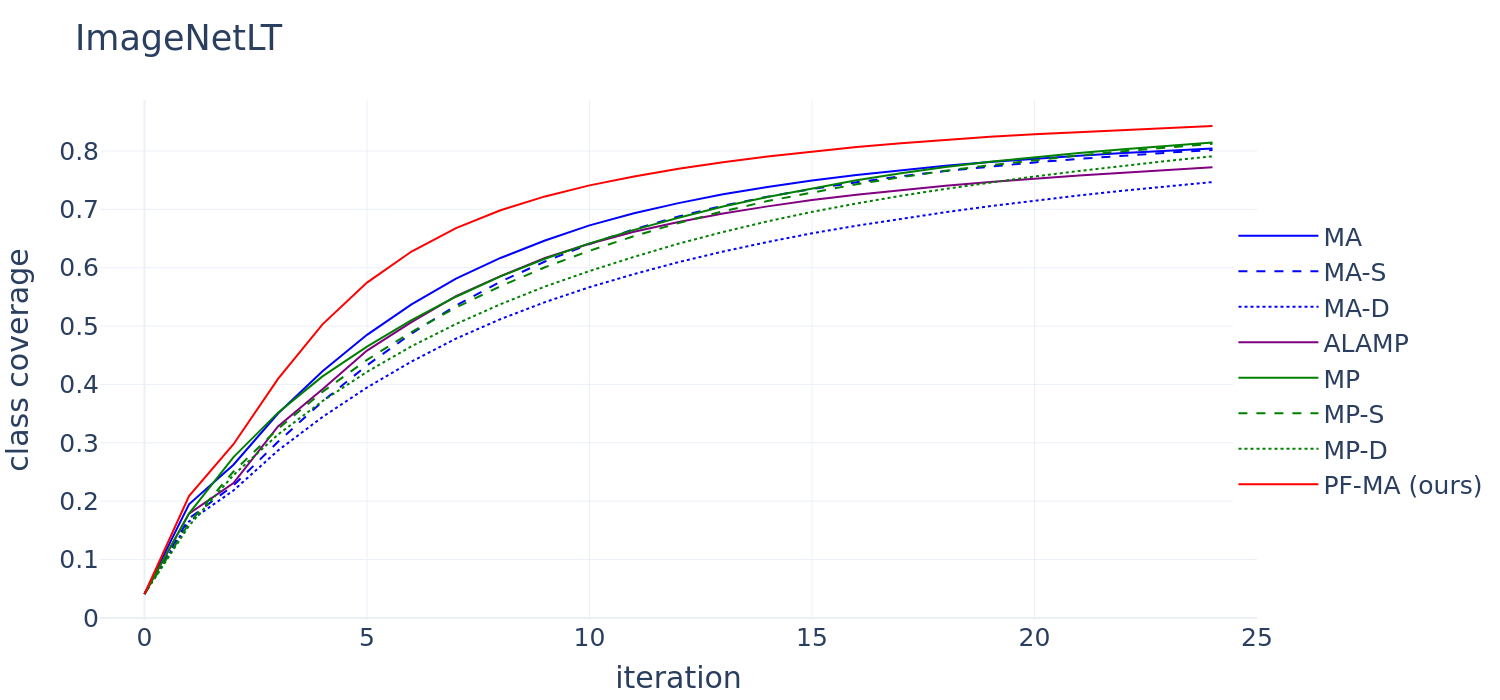}
    \end{subfigure}
    \caption{Class coverage results per iteration.}
    \label{fig:per_iteration}
\end{figure}

Fig. \ref{fig:per_iteration} shows the average retrieval results, for all classes, per iteration, on ImageNet-LT. Additional figures are provided in App.~\ref{app:iteration_class_size}. It is clear that \textit{PF-MA} outperforms all other baselines, with a pronounced improvement in the first few iterations, which is the regime most relevant for interactive human-in-the-loop systems with limited annotation budgets. Its advantage is particularly visible on both Cifar100-LT and ImageNet-LT. We hypothesize that the slight degradation on PlantNet300K is due to the dataset's extreme fine-grained nature, which may require more specialized features. Overall, \textit{PF-MA} reaches a performance level in early iterations that other methods only achieve much later, enabling rapid user satisfaction through relevant, diverse, and rich samples and allowing earlier stopping without sacrificing overall retrieval quality.

\subsubsection{Performance Per Range of Class Size}
\begin{figure}[h!]
    \centering
    \begin{subfigure}[t]{\columnwidth}
        \centering
        \includegraphics[width=\linewidth]{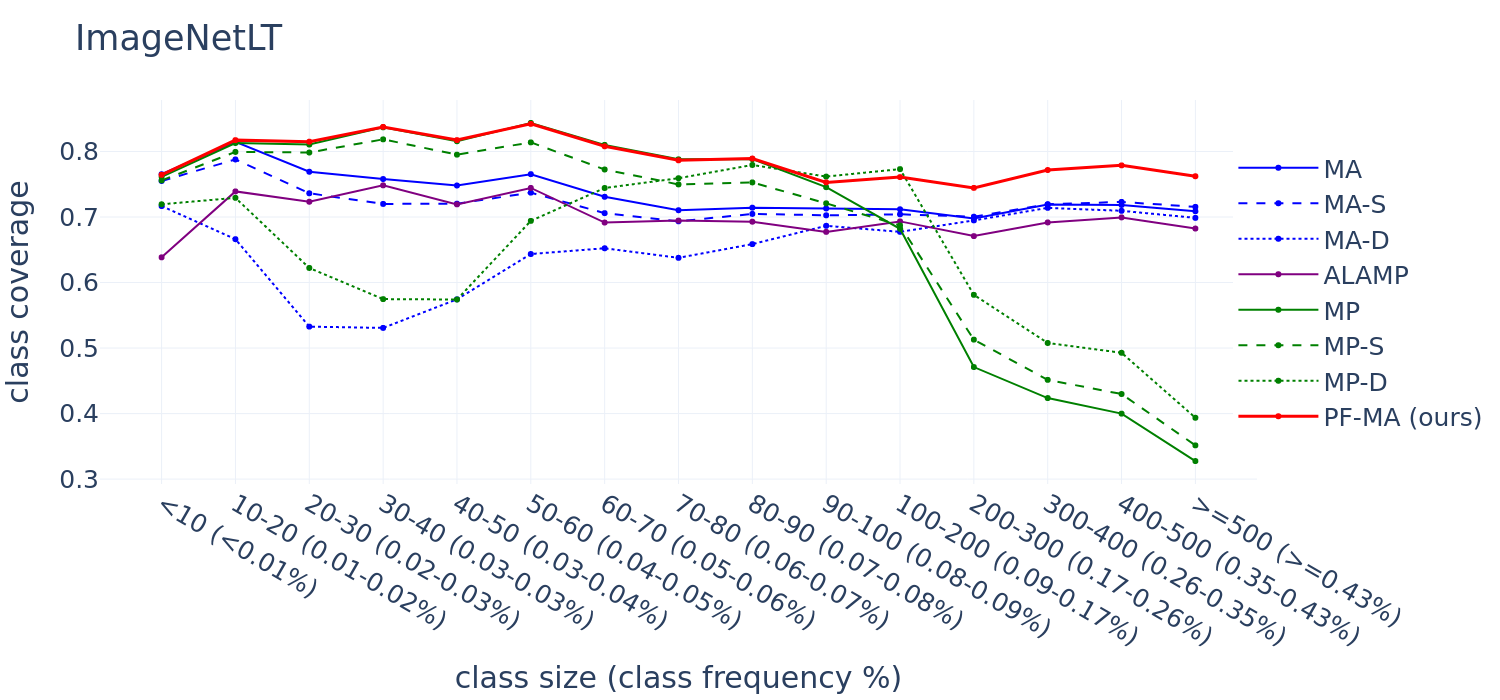}
    \end{subfigure}
    \caption{Class coverage results at iteration $15$ per range of class size.}
    \label{fig:iteration15}
\end{figure}
We analyze the effect of the class-of-interest size on retrieval performance showing results for iteration $15$ in Fig.~\ref{fig:iteration15}. Additional figures are provided in App.~\ref{app:iteration_class_size}. Overall, for small classes, \textit{MP} (and variants) provides better coverage results. As class sizes increase, \textit{MP} suffers a significant drop in performance. Naturally, larger classes present higher probabilities of similar samples, which makes \textit{MP} more prone to return only similar positives. \textit{MA} (and variants) performs better on larger classes. Choosing between \textit{MA} and \textit{MP} would require prior knowledge of class size, which is not available during the interactive retrieval process. \textit{PF-MA}, however, performs consistently well regardless of class size, especially on Cifar100-LT and ImageNet-LT. On PlantNet300K, \textit{MP-D} performs best on medium-sized classes but drops significantly on larger ones, while \textit{MA-D} excels on larger classes but fails on smaller ones. In contrast, \textit{PF-MA} consistently performs well across all class sizes, offering a strong compromise and demonstrating robust, balanced retrieval capabilities. Thus, \textit{PF-MA} can be applied without any prior knowledge regarding the frequency of the desired concept among the whole dataset, eliminating the need to choose between stratgies.

\subsection{Consistent Positive Ratio: Balancing Relevance and Informativeness}
\begin{figure}[h]
    \centering
    \begin{subfigure}[t]{0.49\columnwidth}
        \centering
        \includegraphics[width=\linewidth]{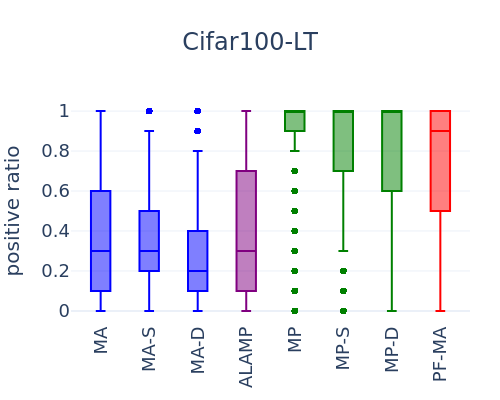}
    \end{subfigure}
    \begin{subfigure}[t]{0.49\columnwidth}
        \centering
        \includegraphics[width=\linewidth]{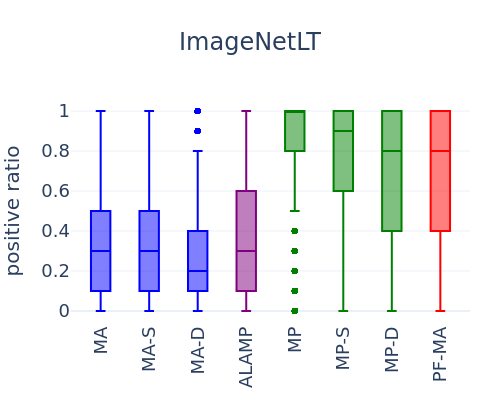}
    \end{subfigure}
    \begin{subfigure}[t]{0.49\columnwidth}
        \centering
        \includegraphics[width=\linewidth]{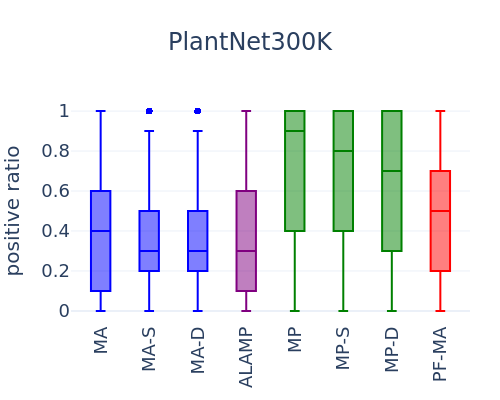}
    \end{subfigure}
    \caption{Distribution of positive ratio among selected samples for different selection strategies.}
    \label{fig:boxplots}
\end{figure}
One of the key advantages of \textit{PF-MA} is its ability to consistently return batches with a high proportion of positive samples across iterations. Fig.~\ref{fig:boxplots} shows the distribution of the percentage of positive samples across all iterations, for different AL strategies, using DINOv2. On Cifar100-LT and ImageNetLT, \textit{PF-MA} is able to maintain consistently high positive rates (more than $80\%$ of the selected batch is positive) with low variance across iterations. This translates to selecting mainly relevant samples, and a few informative negatives. \textit{MA} and its variants, while providing acceptable and competitive performances, as shown in previous sections, mainly return negative samples. This largely impacts user satisfaction while annotating the selected samples, since the majority are irrelevant to the target class. \textit{MP} and its variants, on the other hand, are also able to return a high proportion of positives. However, these strategies consistently underperform \textit{PF-MA} in overall retrieval effectiveness, as well as classifier performance. This highlights that the quality and informativeness of \textit{PF-MA}'s selected positives are superior to simply selecting the most confident positive predictions. Moreover, they sometimes lack negative samples, which naturally hurts the classifier performance. On PlantNet300K, \textit{PF-MA} shows lower overall positive rates due to the challenging nature of the dataset, but still substantially higher than other competitive methods (\textit{MA} and variants). This demonstrates that PF-MA's strategic selection of boundary-proximate positive samples provides the optimal balance between immediate user satisfaction through relevant results and long-term classifier improvement through informative examples.

\section{Conclusion}
In this work, we introduced Positive-First Most Ambiguous \textit{PF-MA}, a simple and efficient active learning strategy tailored to interactive retrieval under severe class imbalance. By explicitly prioritizing boundary-adjacent positive samples, \textit{PF-MA} enables rapid and diverse discovery of rare and visually subtle categories while maintaining strong classifier performance. Our experiments on different long-tailed datasets and pretrained visual representations demonstrate that \textit{PF-MA} consistently outperforms strong baselines in terms of coverage, relevance, and stability. Importantly, our strategy adapts seamlessly across class sizes and does not require prior knowledge of the target class size, making it a practical tool for real-world fine-grained retrieval scenarios. Moreover, we consistently maintain high positive rates, ensuring stable user satisfaction and reliable retrieval performance. Our newly introduced class coverage metric provides a robust measure of retrieval diversity, which is particularly important for fine-grained recognition where multiple visual modes must be explored. 

Overall, our results highlight that aligning active learning with the objectives of user-centric, fine-grained discovery can lead to simple yet effective solutions in realistic human-in-the-loop settings. Future work will explore multimodal extensions combining visual and textual feedback, as well as applications to large-scale biodiversity monitoring and rare species discovery. We also aim to integrate \textit{PF-MA} within foundation model pipelines and interactive systems for real-world fine-grained exploration.

{
    \small
    \bibliographystyle{ieeenat_fullname}
    \bibliography{main}
}

\clearpage
\clearpage
\setcounter{page}{1}
\maketitlesupplementary

\section{Performance of Representativeness-based AL Strategies Across Iterations}
Tab.~\ref{tab:app_coreset} shows the performance of representative-based AL methods for different iterations. The strong performance of the uncertainty-based sampling strategy is consistent from the early retrieval stage.
\label{app:representativeness}
\begin{table}[h]
    \centering
    \caption{Performance comparison of representativeness-based AL methods vs. uncertainty-based selection on CIFAR100-LT at iterations $5$, $15$ and $25$.}
    \label{tab:app_coreset}
    \resizebox{0.75\columnwidth}{!}{
    \begin{tabular}{|c|cccc|}
    \hline
    metric & \textit{Rand} & \textit{DAL} & \textit{CoreSet} & \textit{MA} \\
    \hline
    \hline
    $cov_{5}$ & 0.05 & 0.039 & 0.039 & \textbf{0.504} \\
    $f1_{5}$ & 0.182 & 0.102 & 0.07 & \textbf{0.828} \\
    $pos_{5}$ & 0.022 & 0.019 & 0.02 & \textbf{0.3} \\
    \hline
    \hline
    $cov_{15}$ & 0.081 & 0.066 & 0.058 & \textbf{0.839} \\
    $f1_{15}$ & 0.294 & 0.215 & 0.099 & \textbf{0.857} \\
    $pos_{15}$ & 0.029 & 0.024 & 0.026 & \textbf{0.537} \\
    \hline
    \hline
    $cov_{25}$ & 0.111 & 0.096 & 0.087 & \textbf{0.899} \\
    $f1_{25}$ & 0.381 & 0.321 & 0.164 & \textbf{0.857} \\
    $pos_{25}$ & 0.036 & 0.032 & 0.035 & \textbf{0.61} \\
    \hline
    \end{tabular}
    }
\end{table}

\section{Influence of Coverage Granularity over Iterations}
\label{app:coverage}
Tab.~\ref{tab:app_granularity} extends the results of the effect of coverage granularity on each method's performance to earlier iterations. The results are obtained on ImageNet-LT, using DINOv2 features. Our methods consistently shows strong and stable results since the early retrieval stages, rendering it robust to the choice of $K$. In contrast, other methods are sensitive to this value.
\begin{table}[h!]
    \centering
    \caption{Class coverage scores at iterations $5$, $15$ and $25$ for different $K$ values. Second best is underlined.}
    \label{tab:app_granularity}
    \resizebox{0.75\columnwidth}{!}{
    \begin{tabular}{|c|c|ccc|}
    \hline
    \multirow{ 2}{*}{metric} & \multirow{ 2}{*}{method} & \multicolumn{3}{c|}{$K$} \\ 
    & & 16 & 32 & 64 \\
    \hline
    \hline
    \multirow{8}{*}{$cov_5$}
    & \textit{MA} & \underline{0.55} & \underline{0.423} & 0.327 \\
    & \textit{MP} & 0.413 & 0.396 & \underline{0.399} \\
    & \textit{ALAMP} & 0.49 & 0.387 & 0.315 \\
    & \textit{MA-S} & 0.51 & 0.377 & 0.28 \\
    & \textit{MA-D} & 0.513 & 0.355 & 0.246 \\
    & \textit{MP-S} & 0.427 & 0.381 & 0.343 \\
    & \textit{MP-D} & 0.447 & 0.378 & 0.297 \\
    & \textit{\cellcolor{verylightgray}PF-MA} & \textbf{\cellcolor{verylightgray}0.586} & \textbf{\cellcolor{verylightgray}0.493} & \textbf{\cellcolor{verylightgray}0.4} \\
    \hline
    \hline
    \multirow{8}{*}{$cov_{15}$}
    & \textit{MA} & \underline{0.807} & \underline{0.736} & 0.644 \\
    & \textit{MP} & 0.69 & 0.688 & \underline{0.679} \\
    & \textit{ALAMP} & 0.757 & 0.7 & 0.622 \\
    & \textit{MA-S} & 0.799 & 0.722 & 0.62 \\
    & \textit{MA-D} & 0.774 & 0.653 & 0.527 \\
    & \textit{MP-S} & 0.688 & 0.681 & 0.666 \\
    & \textit{MP-D} & 0.68 & 0.647 & 0.582 \\
    & \textit{\cellcolor{verylightgray}PF-MA} & \textbf{\cellcolor{verylightgray}0.812} & \textbf{\cellcolor{verylightgray}0.79} & \textbf{\cellcolor{verylightgray}0.778} \\
    \hline
    \hline
    \multirow{8}{*}{$cov_{25}$}
    & \textit{MA} & \underline{0.848} & \underline{0.804} & 0.738 \\
    & \textit{MP} & 0.776 & 0.776 & \underline{0.764} \\
    & \textit{ALAMP} & 0.808 & 0.771 & 0.712 \\
    & \textit{MA-S} & \underline{0.848} & 0.802 & 0.731 \\
    & \textit{MA-D} & 0.833 & 0.751 & 0.648 \\
    & \textit{MP-S} & 0.779 & 0.777 & \underline{0.764} \\
    & \textit{MP-D} & 0.773 & 0.76 & 0.723 \\
    & \textit{\cellcolor{verylightgray}PF-MA} & \textbf{\cellcolor{verylightgray}0.849} & \textbf{\cellcolor{verylightgray}0.844} & \textbf{\cellcolor{verylightgray}0.843} \\
    \hline
    \end{tabular}
    }
\end{table}

\section{Performance Across Multiple Datasets and Feature Descriptors}

\subsection{Class coverage scores}
\label{app:generalization_cov}
We report the class coverage scores of the different methods, datasets and feature extractors in Tab.~\ref{tab:app_generalization_imbalanced}, across different iterations. The results highlight consistently strong performance of \textit{PF-MA} since the early stages of retrieval. Other methods do not keep consistent rankings. This further guarantees that our method promotes very early user satisfaction. 
\begin{table*}[h!]
    \centering
    \caption{Class coverage scores at iterations $5$, $15$ and $25$ for different AL methods, datasets and descriptors. Second best is underlined.}
    \label{tab:app_generalization_imbalanced}
    \resizebox{1.3\columnwidth}{!}{
    \begin{tabular}{|c|c|cc|cc|cc|}
    \hline
    \multirow{ 2}{*}{metric} & \multirow{ 2}{*}{method} & \multicolumn{2}{c|}{Cifar100-LT} & \multicolumn{2}{c|}{ImageNet-LT} & \multicolumn{2}{c|}{PlantNet300K}\\ 
    & & CLIP & DINOv2 & CLIP & DINOv2 & CLIP & DINOv2 \\
    \hline
    \hline
    \multirow{8}{*}{$cov_5$}
    & \textit{MA} & 0.402 & \underline{0.502} & 0.351 & \underline{0.423}& 0.203 & 0.282 \\
    & \textit{MP} & \underline{0.41} & 0.454 & \textbf{0.368} & 0.396 & \textbf{0.212} & 0.29 \\
    & \textit{ALAMP} & 0.349 & 0.471 & 0.289 & 0.387 & 0.148 & 0.235 \\
    & \textit{MA-S} & 0.363 & 0.442 & 0.321 & 0.377 & 0.188 & 0.257 \\
    & \textit{MA-D} & 0.388 & 0.407 & 0.327 & 0.355 & 0.193 & 0.284 \\
    & \textit{MP-S} & 0.381 & 0.435 & 0.34 & 0.381 & 0.193 & 0.27\\
    & \textit{MP-D} & 0.409 & 0.429 & 0.347 & 0.378 & 0.198 & \underline{0.294}\\
    & \textit{\cellcolor{verylightgray}PF-MA} & \textbf{\cellcolor{verylightgray}0.411} & \textbf{\cellcolor{verylightgray}0.56} & \underline{\cellcolor{verylightgray}0.36} & \textbf{\cellcolor{verylightgray}0.493} & \underline{\cellcolor{verylightgray}0.204}& \textbf{\cellcolor{verylightgray}0.298} \\
    \hline
    \hline
    \multirow{8}{*}{$cov_{15}$} 
    & \textit{MA} & 0.801 & \underline{0.838}& \underline{0.76}& \underline{0.736}& \underline{0.487}& \underline{0.59}\\
    & \textit{MP} & 0.773 & 0.785 & 0.713 & 0.688 & 0.456 & 0.537 \\
    & \textit{ALAMP} & 0.704 & 0.83 & 0.638 & 0.7 & 0.318 & 0.514 \\
    & \textit{MA-S} & 0.775 & 0.821 & 0.735 & 0.722 & 0.465 & 0.57 \\
    & \textit{MA-D} & \underline{0.815}& 0.74 & 0.734 & 0.653 & 0.48 & 0.588 \\
    & \textit{MP-S} & 0.756 & 0.785 & 0.694 & 0.681 & 0.435 & 0.523 \\
    & \textit{MP-D} & 0.774 & 0.736 & 0.697 & 0.647 & 0.447 & 0.542 \\
    & \textit{\cellcolor{verylightgray}PF-MA} & \textbf{\cellcolor{verylightgray}0.824} & \textbf{\cellcolor{verylightgray}0.915} & \textbf{\cellcolor{verylightgray}0.773} & \textbf{\cellcolor{verylightgray}0.79} & \textbf{\cellcolor{verylightgray}0.488} & \textbf{\cellcolor{verylightgray}0.604} \\
    \hline
    \hline
    \multirow{8}{*}{$cov_{25}$} 
    & \textit{MA} & 0.89 & \underline{0.899}& \underline{0.852}& \underline{0.804}& \textbf{0.596} & 0.678 \\
    & \textit{MP} & 0.875 & 0.877 & 0.821 & 0.776 & 0.565 & 0.628 \\
    & \textit{ALAMP} & 0.811 & 0.892 & 0.744 & 0.771 & 0.406 & 0.605 \\
    & \textit{MA-S} & 0.88 & 0.895 & 0.844 & 0.802 & 0.583 & 0.671 \\
    & \textit{MA-D} & \underline{0.903}& 0.84 & 0.842 & 0.751 & \underline{0.595}& \underline{0.679}\\
    & \textit{MP-S} & 0.869 & 0.882 & 0.815 & 0.777 & 0.552 & 0.625 \\
    & \textit{MP-D} & 0.881 & 0.86 & 0.816 & 0.76 & 0.563 & 0.639 \\
    & \textit{\cellcolor{verylightgray}PF-MA} & \textbf{\cellcolor{verylightgray}0.908} & \textbf{\cellcolor{verylightgray}0.954} & \textbf{\cellcolor{verylightgray}0.861} & \textbf{\cellcolor{verylightgray}0.844} & \textbf{\cellcolor{verylightgray}0.596} & \textbf{\cellcolor{verylightgray}0.684} \\
    \hline
    \end{tabular}
    }
\end{table*} 

\subsection{Classifier performance}
\label{app:generalization_f1}
\begin{table*}[h!]
    \centering
    \caption{F1 scores at iterations $5$, $15$ and $25$ for different AL methods, datasets and descriptors. Top-3 results are underlined. A star (*) is used only when our method achieves the top-1 result.}
    \label{tab:app_generalization_imbalanced_f1}
    \resizebox{1.3\columnwidth}{!}{
    \begin{tabular}{|c|c|cc|cc|cc|}
    \hline
    \multirow{ 2}{*}{metric} & \multirow{ 2}{*}{method} & \multicolumn{2}{c|}{Cifar100-LT} & \multicolumn{2}{c|}{ImageNet-LT} & \multicolumn{2}{c|}{PlantNet300K}\\ 
    & & CLIP & DINOv2 & CLIP & DINOv2 & CLIP & DINOv2 \\
    \hline
    \hline
    \multirow{8}{*}{$f1_5$}
    & \textit{MA} & \underline{0.604} & \underline{0.83} & \underline{0.504} & \underline{0.675} & \underline{0.191} & \underline{0.413} \\
    & \textit{MP} & 0.258 & 0.597 & 0.187 & 0.573 & 0.077 & 0.221 \\
    & \textit{ALAMP} & 0.56 & 0.806 & 0.45 & 0.646 & 0.147 & 0.373 \\
    & \textit{MA-S} & \underline{0.57} & \underline{0.816} & \underline{0.476} & \underline{0.667} & \underline{0.176} & \underline{0.398} \\
    & \textit{MA-D} & 0.517 & 0.803 & 0.437 & 0.661 & \underline{0.159} & \underline{0.398} \\
    & \textit{MP-S} & 0.325 & 0.666 & 0.241 & 0.59 & 0.091 & 0.279 \\
    & \textit{MP-D} & 0.358 & 0.637 & 0.245 & 0.579 & 0.093 & 0.296 \\
    & \textit{\cellcolor{verylightgray}PF-MA} & \underline{\cellcolor{verylightgray}0.601} & \underline{\cellcolor{verylightgray}0.814} & \underline{\cellcolor{verylightgray}0.505}* & \underline{\cellcolor{verylightgray}0.67} & \underline{\cellcolor{verylightgray}0.191}* & \underline{\cellcolor{verylightgray}0.41} \\
    \hline
    \hline
    \multirow{8}{*}{$f1_{15}$} 
    & \textit{MA} & \underline{0.659} & \underline{0.86} & \underline{0.59} & \underline{0.727} & \underline{0.296} & \underline{0.498} \\
    & \textit{MP} & 0.431 & 0.71 & 0.378 & 0.63 & 0.174 & 0.388 \\
    & \textit{ALAMP} & 0.644 & \underline{0.858} & 0.56 & 0.709 & 0.237 & 0.476 \\
    & \textit{MA-S} & \underline{0.656} & \underline{0.858} & \underline{0.584} & \underline{0.725} & 0.283 & \underline{0.493} \\
    & \textit{MA-D} & 0.637 & 0.85 & 0.571 & \underline{0.723} & \underline{0.268} & 0.492 \\
    & \textit{MP-S} & 0.445 & 0.727 & 0.387 & 0.631 & 0.174 & 0.39 \\
    & \textit{MP-D} & 0.438 & 0.725 & 0.383 & 0.624 & 0.17 & 0.396 \\
    & \textit{\cellcolor{verylightgray}PF-MA} & \underline{\cellcolor{verylightgray}0.665}* & \underline{\cellcolor{verylightgray}0.861}* & \underline{\cellcolor{verylightgray}0.587} & \underline{\cellcolor{verylightgray}0.723} & \underline{\cellcolor{verylightgray}0.296}* & \underline{\cellcolor{verylightgray}0.495} \\
    \hline
    \hline
    \multirow{8}{*}{$f1_{25}$} 
    & \textit{MA} & \underline{0.648} & \underline{0.859} & \underline{0.593} & \underline{0.735} & \underline{0.298} & \underline{0.508} \\
    & \textit{MP} & 0.456 & 0.73 & 0.416 & 0.63 & 0.196 & 0.421 \\
    & \textit{ALAMP} & \underline{0.647} & \underline{0.864} & 0.58 & 0.722 & 0.266 & 0.493 \\
    & \textit{MA-S} & \underline{0.649} & \underline{0.862} & \underline{0.593} & \underline{0.735} & \underline{0.295} & \underline{0.506} \\
    & \textit{MA-D} & \underline{0.648} & 0.856 & \underline{0.589} & \underline{0.734} & 0.29 & \underline{0.508} \\
    & \textit{MP-S} & 0.472 & 0.746 & 0.427 & 0.632 & 0.202 & 0.429 \\
    & \textit{MP-D} & 0.49 & 0.742 & 0.434 & 0.632 & 0.207 & 0.434 \\
    & \textit{\cellcolor{verylightgray}PF-MA} & \underline{\cellcolor{verylightgray}0.649}* & \underline{\cellcolor{verylightgray}0.859} & \underline{\cellcolor{verylightgray}0.583} & \underline{\cellcolor{verylightgray}0.73} & \underline{\cellcolor{verylightgray}0.299}* & \underline{\cellcolor{verylightgray}0.507} \\
    \hline
    \end{tabular}
    }
\end{table*}
Tab.~\ref{tab:app_generalization_imbalanced_f1} reports the f1-score of the classifier across different methods, datasets and feature extractors. \textit{PF-MA} is always among the top-3 methods in terms of classifier generalization, with minimal degradation, although our main objective is retrieval. Other competing methods have much lower coverage results as discussed before. Thus, \textit{PF-MA} is able to maintain a fairly generalizable classifier, while achieving better coverage results.

\section{Performance Across the Search Iterations and Per range of Class Size}

\subsection{Class coverage scores}
\label{app:iteration_class_size}
\begin{figure}[h!]
    \centering
    \begin{subfigure}[t]{\columnwidth}
        \centering
        \includegraphics[width=\linewidth]{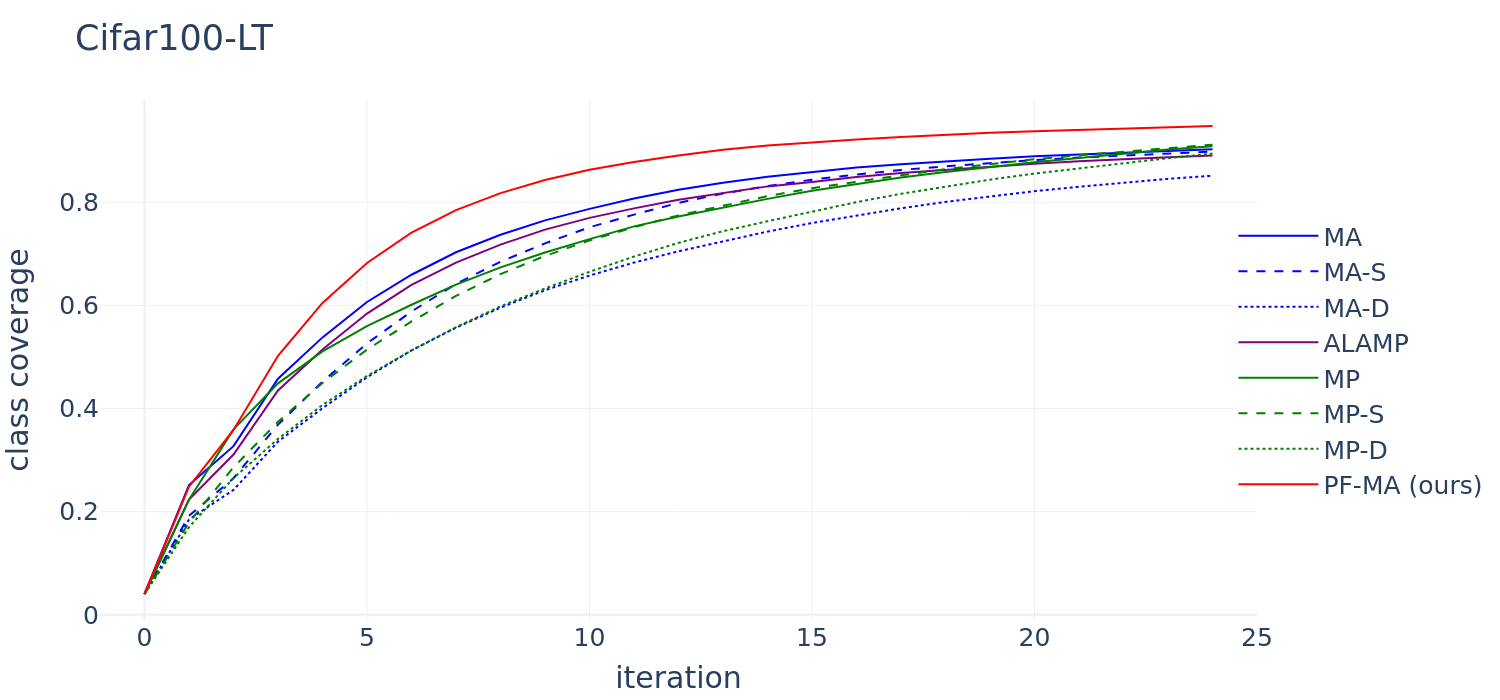}
    \end{subfigure}
    \begin{subfigure}[t]{\columnwidth}
        \centering
        \includegraphics[width=\linewidth]{figures/FIG_imagenetlt_class_coverage_per_iteration.png}
    \end{subfigure}
    \begin{subfigure}[t]{\columnwidth}
        \centering
        \includegraphics[width=\linewidth]{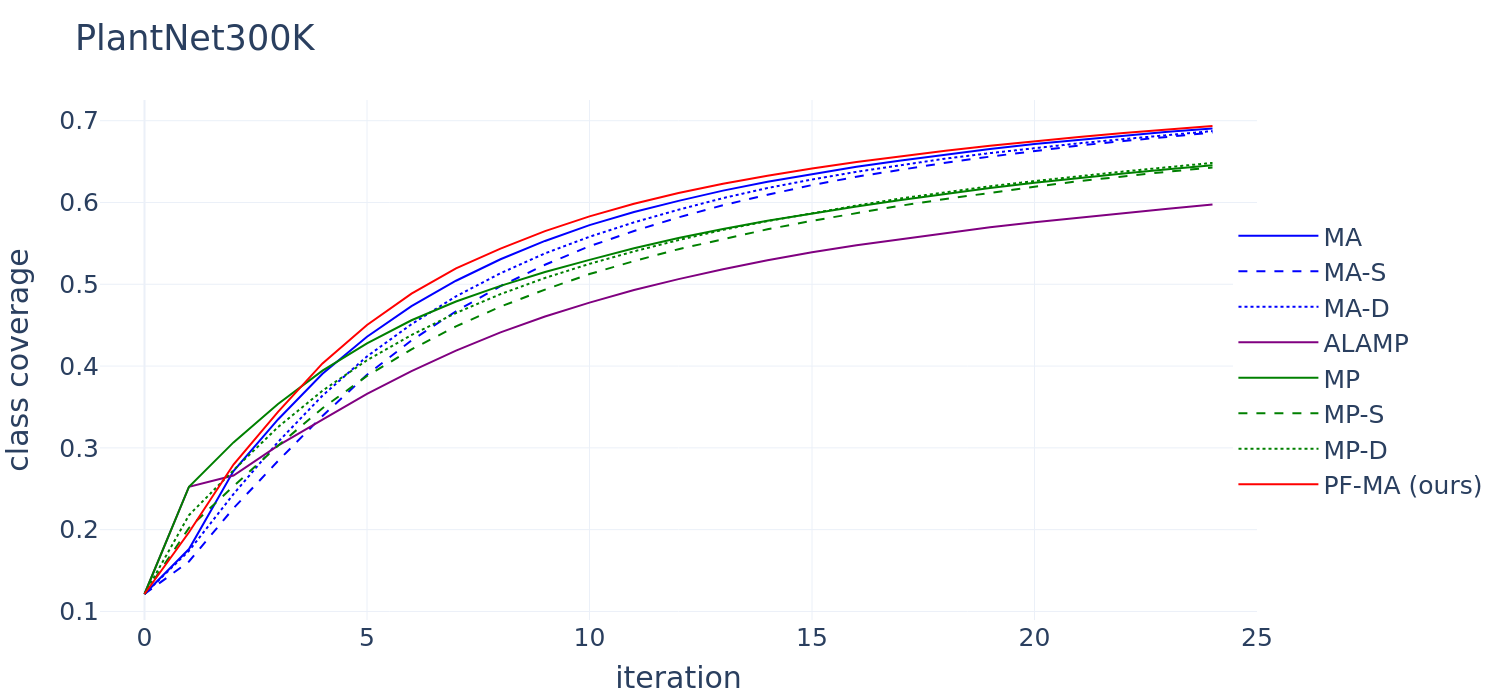}
    \end{subfigure}
    \caption{Class coverage results per iteration.}
    \label{fig:app_per_iteration}
\end{figure}

\begin{figure}[h!]
    \centering
    \begin{subfigure}[t]{\columnwidth}
        \centering
        \includegraphics[width=\linewidth]{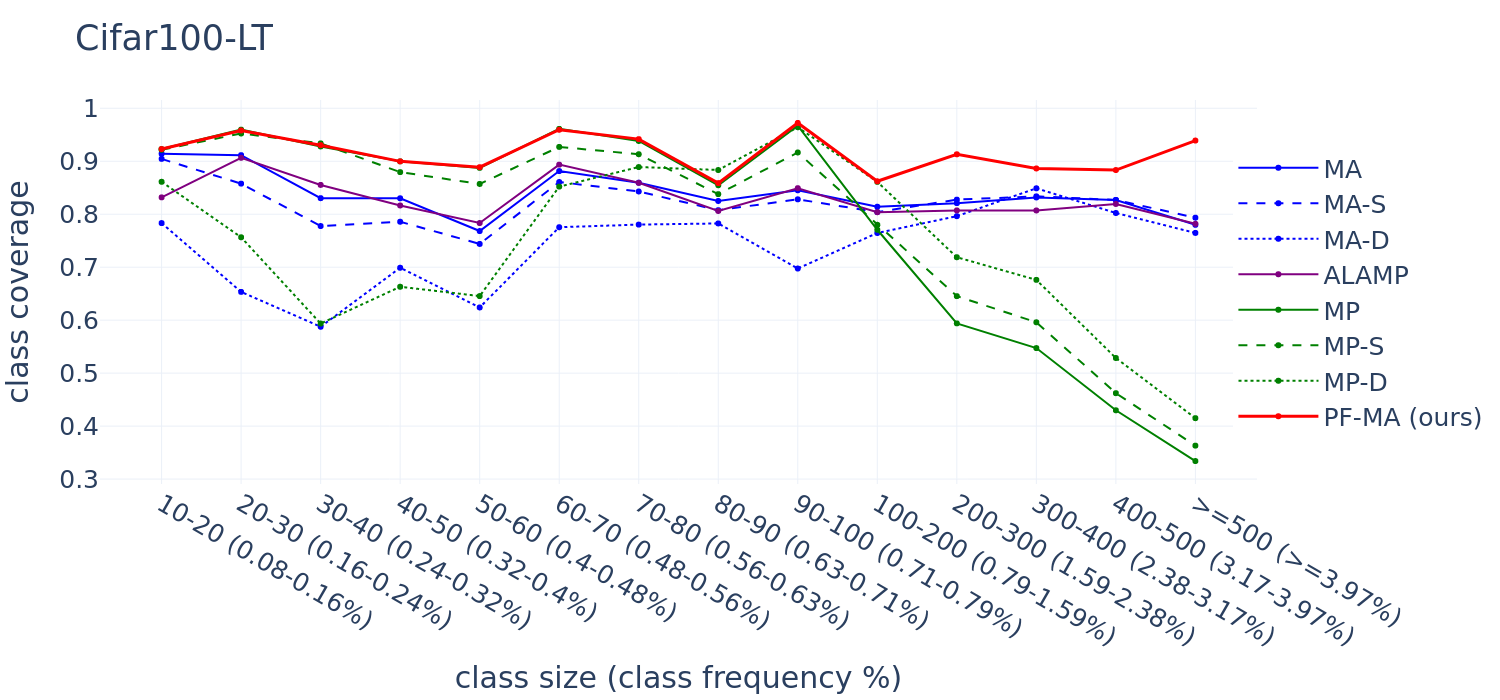}
    \end{subfigure}
    \begin{subfigure}[t]{\columnwidth}
        \centering
        \includegraphics[width=\linewidth]{figures/FIG_imagenetlt_class_coverage_iteration15.png}
    \end{subfigure}
    \begin{subfigure}[t]{\columnwidth}
        \centering
        \includegraphics[width=\linewidth]{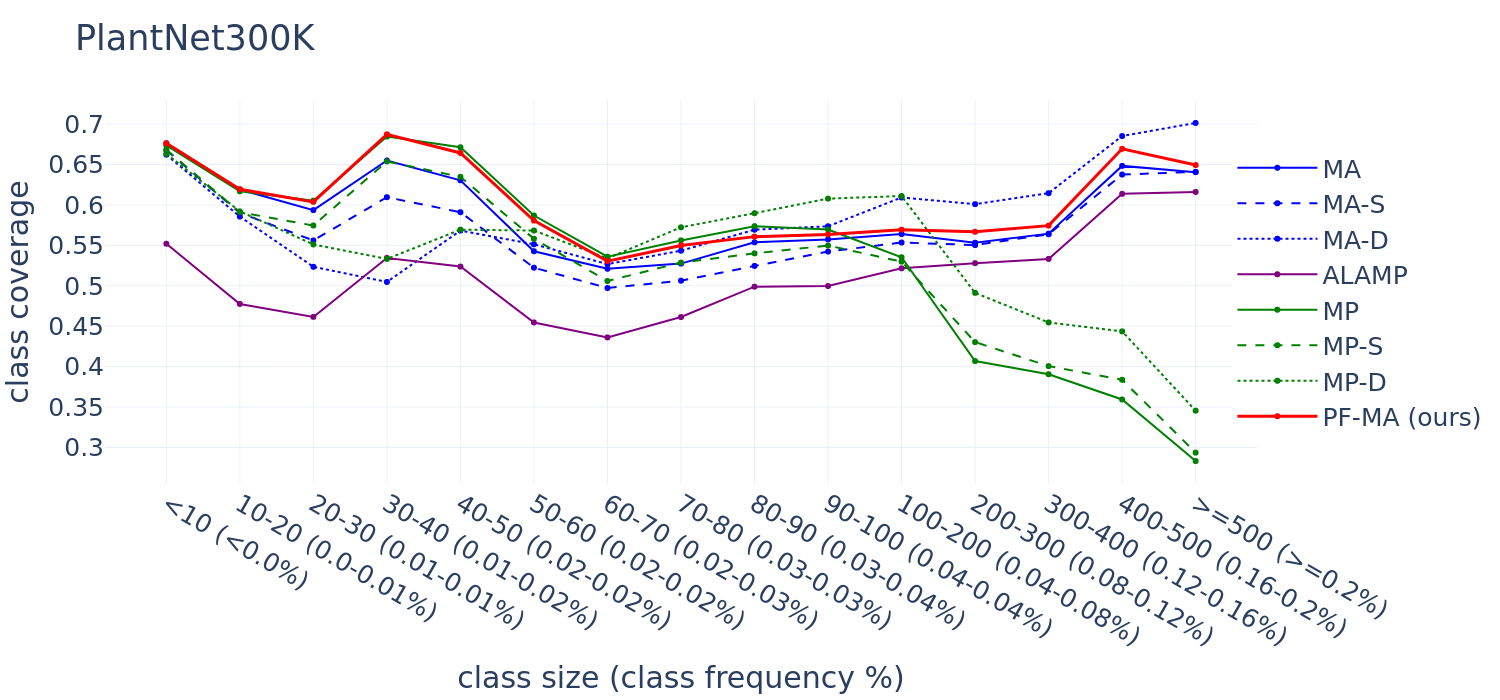}
    \end{subfigure}
    \caption{Class coverage results at iteration $15$ per range of class size.}
    \label{fig:app_iteration15}
\end{figure}

Fig.~\ref{fig:app_per_iteration} and Fig.~\ref{fig:app_iteration15} show the strong performance of \textit{PF-MA} across iterations and per range of class size, for different datasets. The results are less pronounced on PlantNet300K because of its specific nature. \textit{MP} and variants perform well on smaller classes, but they struggle for larger ones, where \textit{MA} perform better thanks to the less imbalanced distribution of the negative vs. positive class problem. Our method yields consistently higher coverage across iterations, and presents the best compromise between smaller and larger classes. This highlights its practicality at test time, where early user satisfaction is desired, and the actual size of the user-defined concept in unknown.

\subsection{Classifier performance}
\label{app:iteration_class_size_f1}
\begin{figure}[h!]
    \centering
    \begin{subfigure}[t]{\columnwidth}
        \centering
        \includegraphics[width=\linewidth]{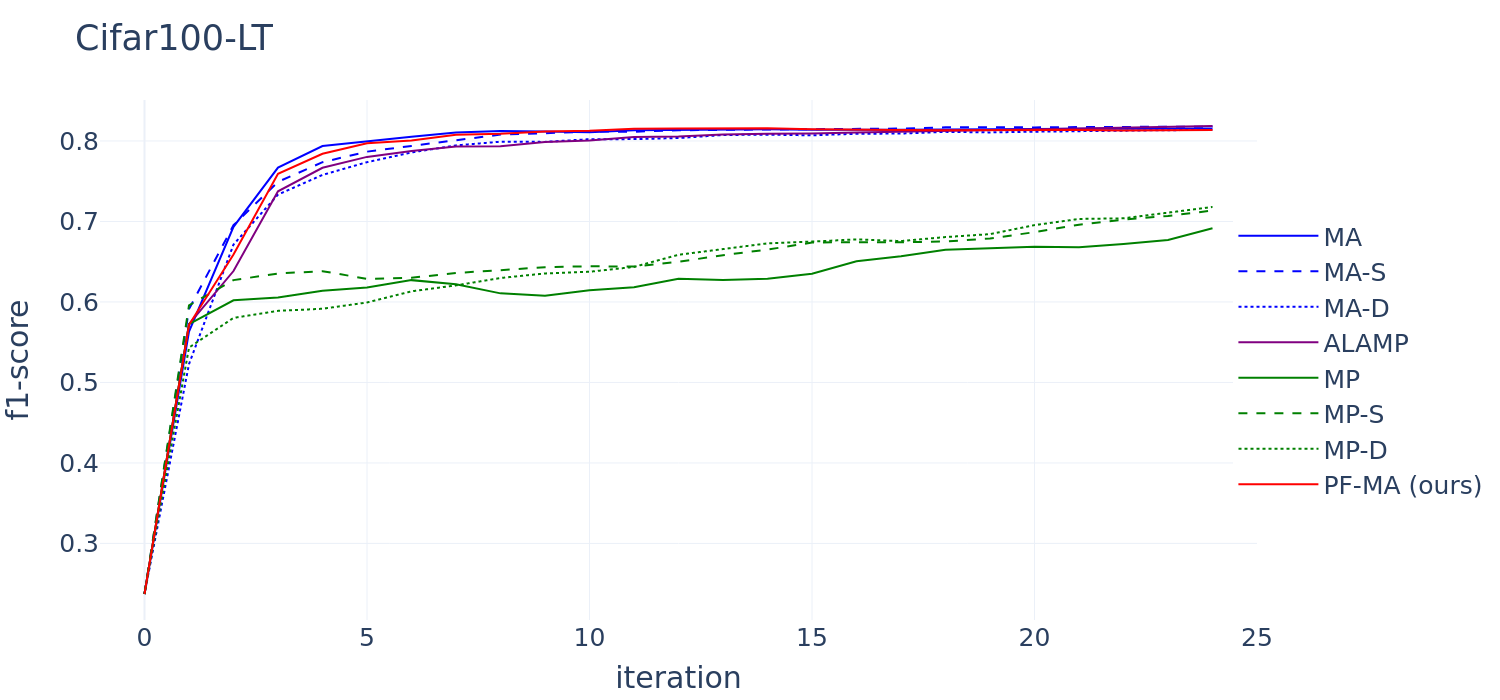}
    \end{subfigure}
    \begin{subfigure}[t]{\columnwidth}
        \centering
        \includegraphics[width=\linewidth]{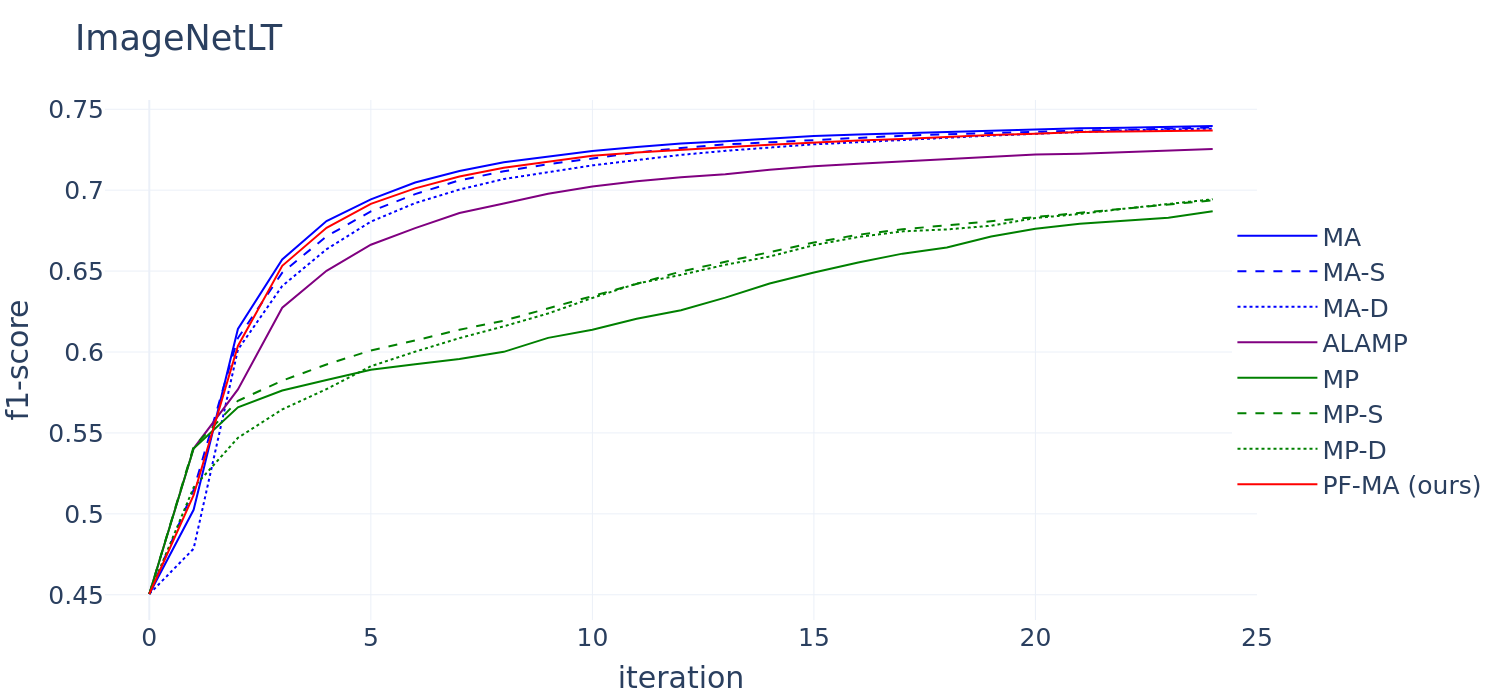}
    \end{subfigure}
    \begin{subfigure}[t]{\columnwidth}
        \centering
        \includegraphics[width=\linewidth]{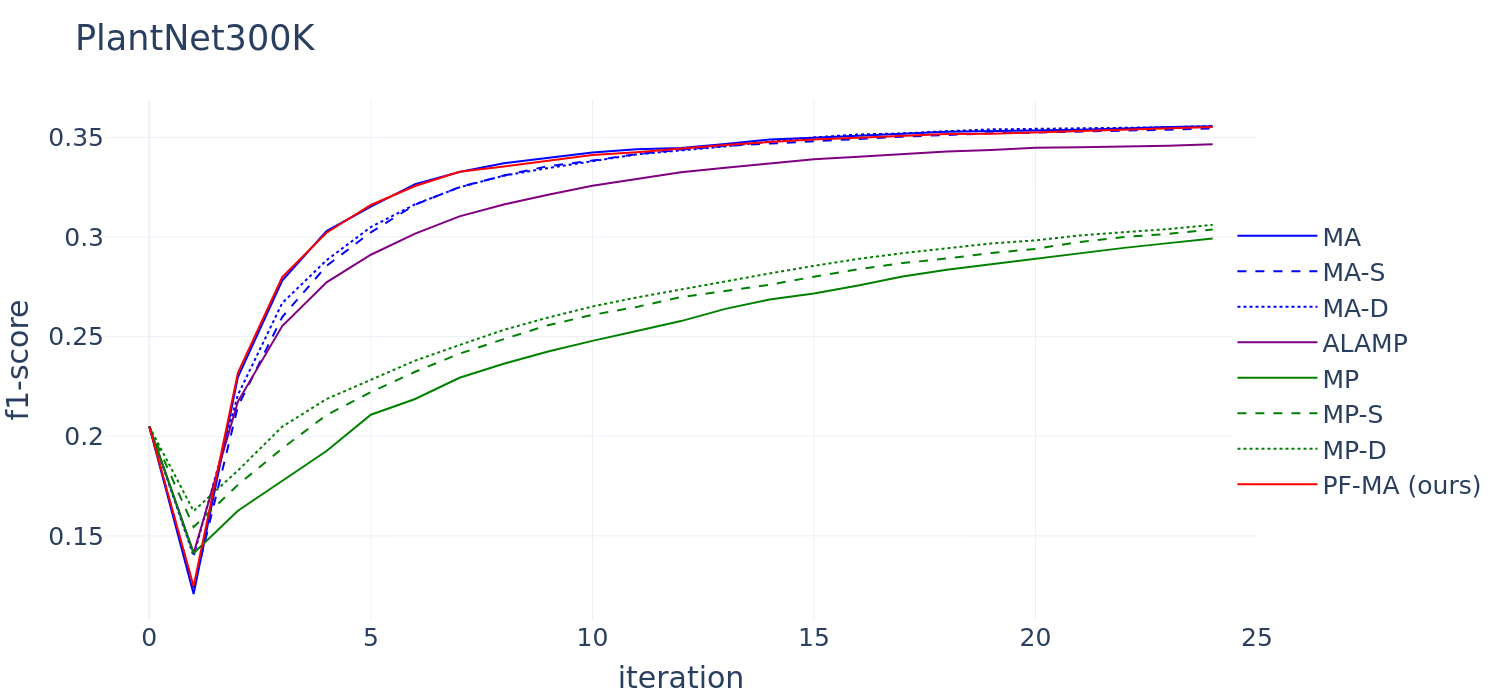}
    \end{subfigure}
    \caption{F1 score results per iteration.}
    \label{fig:per_iteration_f1}
\end{figure}

\begin{figure}[h!]
    \centering
    \begin{subfigure}[t]{\columnwidth}
        \centering
        \includegraphics[width=\linewidth]{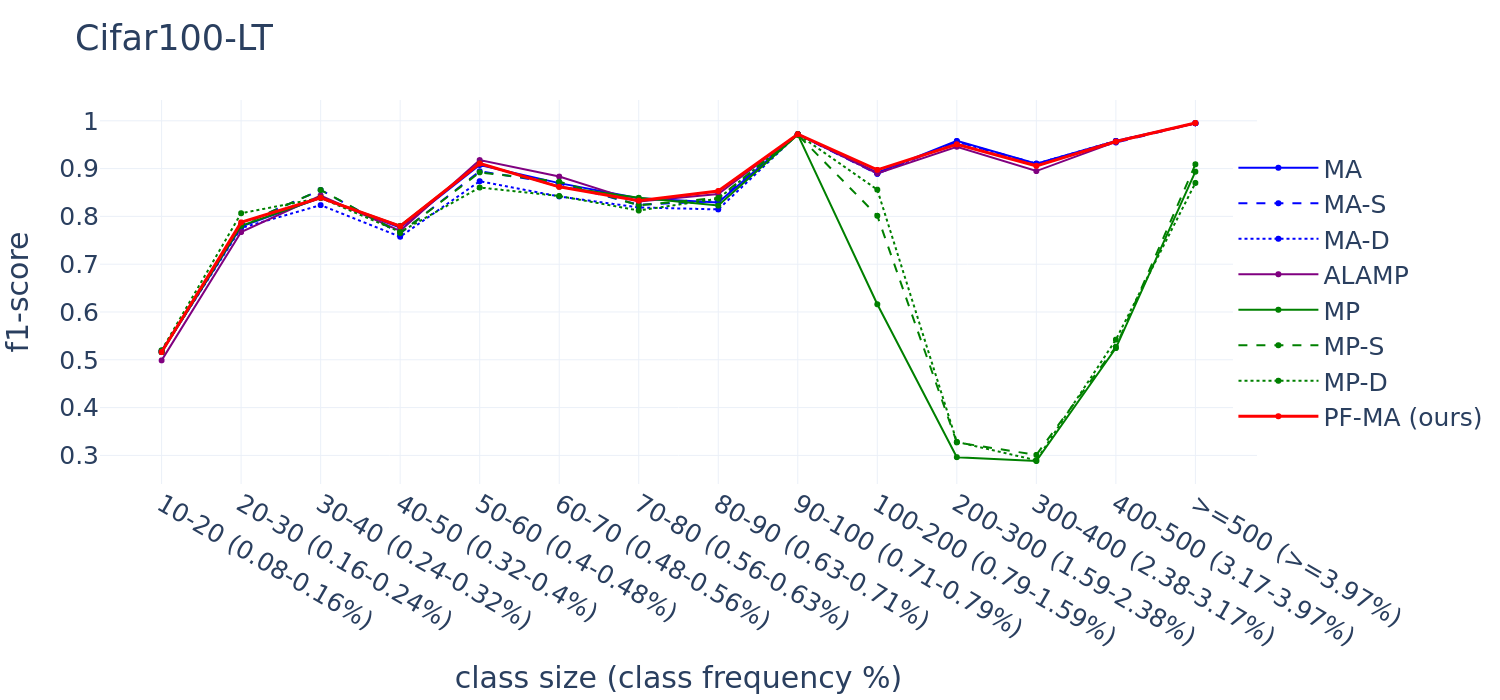}
    \end{subfigure}
    \begin{subfigure}[t]{\columnwidth}
        \centering
        \includegraphics[width=\linewidth]{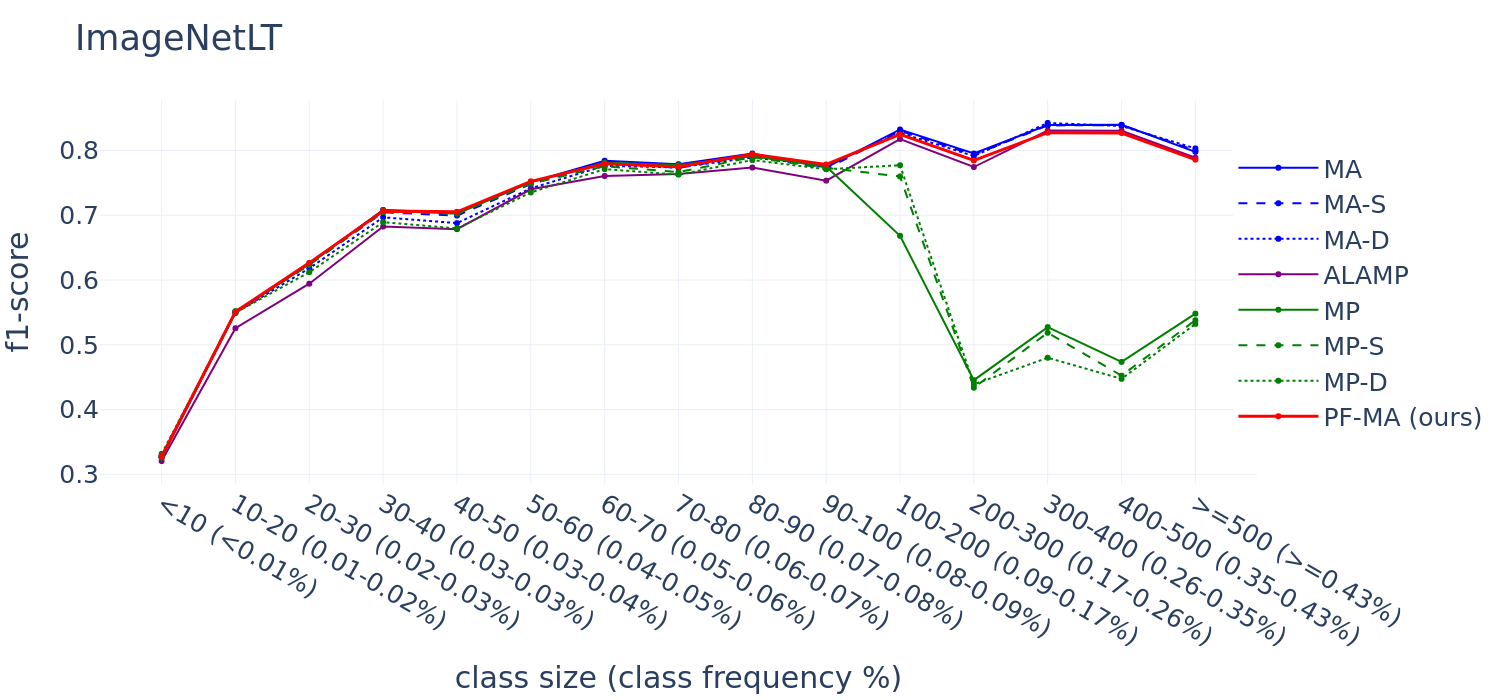}
    \end{subfigure}
    \begin{subfigure}[t]{\columnwidth}
        \centering
        \includegraphics[width=\linewidth]{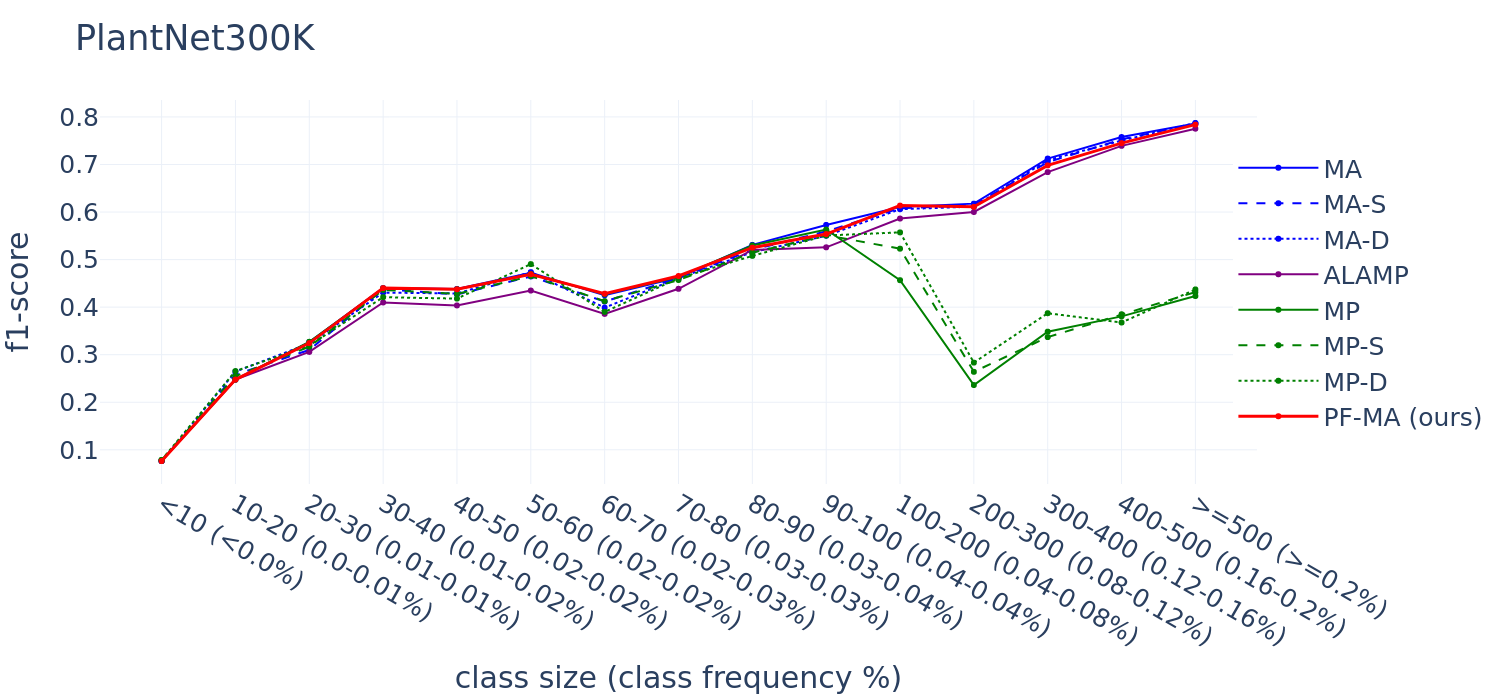}
    \end{subfigure}
    \caption{F1 score results at iteration $15$ per range of class size.}
    \label{fig:iteration15_f1}
\end{figure}

Both Fig.~\ref{fig:per_iteration_f1} and Fig.~\ref{fig:iteration15_f1} show the ability of \textit{PF-MA} to achieve competitive classifier generalization performance. This performance is steady across iterations, and across class sizes. \textit{MP} on the other hand, has very poor classifier performance overall, always with a significant drop on larger class sizes.

\end{document}